\documentclass[journal]{IEEEtran}
\usepackage[cmex10]{amsmath}
\usepackage{amssymb}

\usepackage{cite}

\makeatletter
\def\endthebibliography{%
	\def\@noitemerr{\@latex@warning{Empty `thebibliography' environment}}%
	\endlist
}
\makeatother

\ifCLASSINFOpdf
\else
\fi
\usepackage{amsmath}

\usepackage{float}
\usepackage{graphicx}
\usepackage[T1]{fontenc}
\usepackage{subfig} 
\usepackage[official]{eurosym}
\usepackage{amssymb}

\usepackage{hyperref}

\begin{document}
	%
	\title{Deep Learning Based Video System for \\
		Accurate and Real-Time Parking Measurement}
	%
	%
	%
	
	\author{Bill Yang~Cai,~\IEEEmembership{Student Member,~IEEE,}
		Ricardo~Alvarez, Michelle~Sit, F\'{a}bio~Duarte, and Carlo~Ratti
		\thanks{Bill Y. Cai, Ricardo Alvarez, F\'{a}bio~Duarte, and Carlo Ratti are with the Senseable City Lab, Massachusetts Institute of Technology, 77 Massachusetts Avenue, Cambridge, MA 02139, USA. Bill Y. Cai is also with the Center for Computational Engineering at the Massachusetts Institute of Technology, 77 Massachusetts Avenue, Cambridge, MA 02139, USA. F\'{a}bio~Duarte is also with the School of Architecture and Design at the Pontifical Catholic University of Paran\'{a},  R. Imac. Conceição, 1155 - Prado Velho, Curitiba - PR, 80215-901, Brazil. (email: me@billcai.com, jraf@mit.edu, fduarte@mit.edu, ratti@mit.edu)}
		\thanks{Michelle Sit is with the Department of Computer Science and Engineering at the University of California, San Diego, 9500 Gilman Drive, La Jolla, CA 92093, USA. (email: mcsit@eng.ucsd.edu)}
		\thanks{Manuscript received 29 Aug 2018, First revision submitted 22 Dec 2018, Accepted for publication 18 Feb 2019}}
	
	%
	%

	\markboth{IEEE Internet of Things Journal: Special Issue on
		Enabling a Smart City: Internet of Things Meets AI}%
	{Shell \MakeLowercase{\textit{et al.}}: Bare Demo of IEEEtran.cls for IEEE Journals}
	%



	\maketitle
	
	\begin{abstract}
		Parking spaces are costly to build, parking payments are difficult to enforce, and drivers waste an excessive amount of time searching for empty lots. Accurate quantification would inform developers and municipalities in space allocation and design, while real-time measurements would provide drivers and parking enforcement with information that saves time and resources. In this paper, we propose an accurate and real-time video system for future Internet of Things (IoT) and smart cities applications. Using recent developments in deep convolutional neural networks (DCNNs) and a novel vehicle tracking filter, we combine information across multiple image frames in a video sequence to remove noise introduced by occlusions and detection failures. We demonstrate that our system achieves higher accuracy than pure image-based instance segmentation, and is comparable in performance to industry benchmark systems that utilize more expensive sensors such as radar. Furthermore, our system shows significant potential in its scalability to a city-wide scale and also in the richness of its output that goes beyond traditional binary occupancy statistics.
	\end{abstract}
	
	\begin{IEEEkeywords}
		Internet of Things (IoT), smart city, parking, computer vision, deep learning, artificial intelligence
	\end{IEEEkeywords}

	%
	\IEEEpeerreviewmaketitle

	\section{Introduction}
	%
	%
	%
	%
	\IEEEPARstart{I}{n} cities, parking lots are costly in terms of space, construction and maintenance costs. Parking lots take up 6.57\% of urban land use \cite{davis2010environmental} and collectively make up nearly 7000 $\text{km}^2$ of land use in the United States \cite{jakle2004lots}. In U.S. cities, the area of parking lots take up more than 3 times the area of urban parks \cite{davis2010environmental}. Government-set requirements for private developers\footnote{An example of government-set requirements on parking spaces can be seen in the municipal code of Placer County, CA: \url{https://qcode.us/codes/placercounty/view.php?topic=17-2-vii-17_54-17_54_060}. For example, the municipal code of Placer County states that restaurants are required to have 1 parking spot per 100 square feet of floor area, and shopping centers are required to have 1 parking spot per 200 square feet of floor area.} often require developers to provide parking lots to meet peak parking demand, resulting in an excessive number of parking lots with low utilization  \cite{davis2010environmental}. Construction costs (excluding land acquisition) of an average parking lot costs nearly \$20,000 in the United States, while examples of annual maintenance cost of a single parking lot include \$461 in Fort Collins, CO \cite{VTPIreport} and \$729 in San Francisco, CA \cite{natstreetreport}. 
	
	Drivers also spend a significant amount of time looking for parking, overpay for what they use, and pay a high amount of parking fines. A recent study extrapolates from surveys taken in 10 American cities, 10 British cities, and 10 German cities, and found that drivers in the United States, United Kingdom, and Germany spend averages of 17, 44 and 41 hours annually respectively to find parking\footnote{The parking search time in the United States, United Kingdom, and Germany translates to an economic cost of \$72.7b, \$29.7b, and \$46.2b \cite{inrixreport}.}\cite{inrixreport}. In larger US cities, the search cost is significantly higher, with an estimated 107 hours annually in New York City and 85 hours in Los Angeles. This additional search time also contributes to congestion and air pollution in cities \cite{shoup2006cruising}. The same survey found that drivers overpay fees for 13 to 45 hours of parking per annum, and were also fined \$2.6b, \$1.53b, and \$434m over the period of a year in the United States, United Kingdom and Germany respectively\cite{inrixreport}. 
	
	Cities can benefit tremendously from a large-scale and accurate study of parking behavior. Especially with the possible transformation in land transportation brought on by autonomous vehicles \cite{duarte2018impact}, policymakers and urban planners would benefit from a scalable and accurate measurement and analysis of parking trends and activities. Additionally, real-time parking quantification method can also provide human or computer drivers with relevant parking information to reduce search time and allow for efficient route planning. In addition, a real-time parking measurement system can also enhance parking enforcement in cities; using May 2017 estimates by the Bureau of Labor Statistics, the labor costs in terms of just salary paid to parking enforcement officers amount to more than \$350 million annually \cite{blsreport}. The use of such a technical solution will allow these officers to have "eyes" on the ground, cover more parking lots with less physical effort, and increase municipal revenues from parking fines. A real-time, scalable and accurate parking measurement method is also a critical capability required by future parking system features, such as demand-based pricing \cite{pierce2013getting} and reservation for street parking \cite{geng2013new}, that have the potential to make parking more convenient for drivers, and to incentivize socially beneficial behavior.
	\begin{figure*}
		\centering
		\resizebox{0.623\columnwidth}{!}{%
			\includegraphics[scale=0.5]{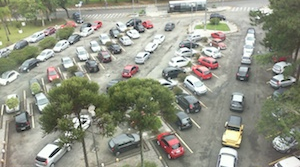}}
		\resizebox{0.623\columnwidth}{!}{%
			\includegraphics[scale=0.5]{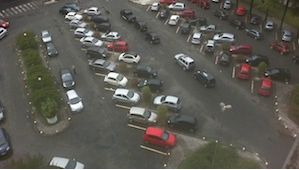}}
		\resizebox{0.623\columnwidth}{!}{%
			\includegraphics[scale=0.5]{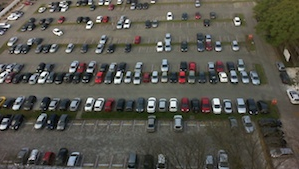}}\\
		\vspace{1mm}
		\resizebox{0.623\columnwidth}{!}{%
			\includegraphics[scale=0.5]{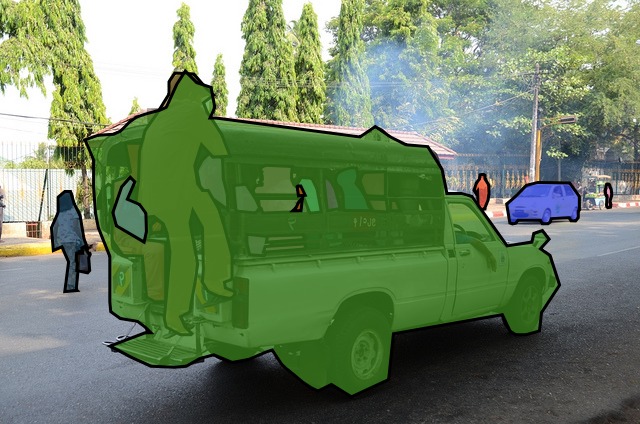}}
		\resizebox{0.623\columnwidth}{!}{%
			\includegraphics[scale=0.5]{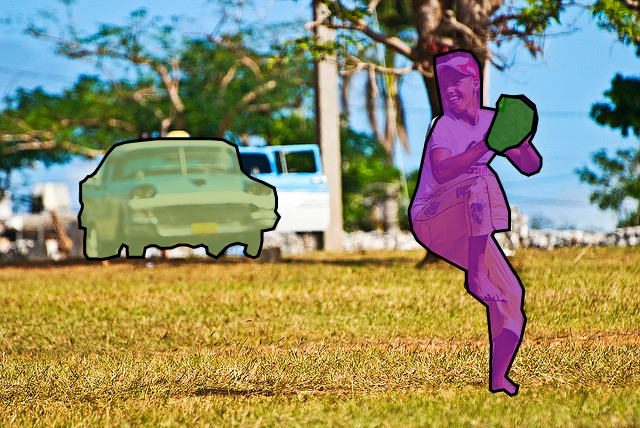}}
		\resizebox{0.623\columnwidth}{!}{%
			\includegraphics[scale=0.5]{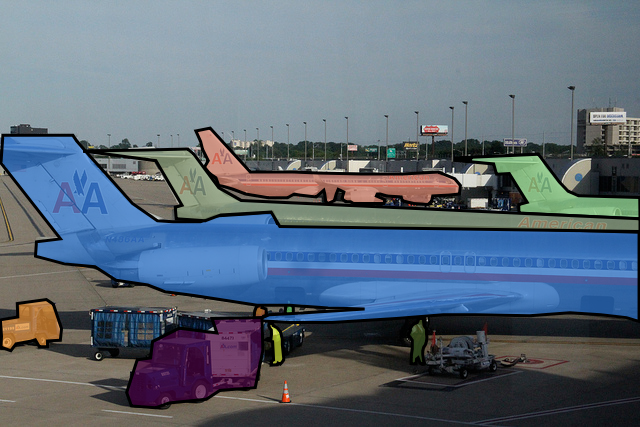}}
		\caption[Sample images from PKLot dataset and Common Objects in Context dataset]{Top 3 images show the 3 different perspectives that the PKLot dataset by Almeida et al \cite{de2015pklot} is obtained from. Bottom 3 images show 3 sample images that is in the COCO dataset, reflecting the diverse contexts that object instances are identified in.}
		\label{fig:cocopklot}
	\end{figure*}
	
	\section{Existing quantification methods}
	Existing parking utilization methods can be divided into three types: counter-based, sensor-based and image-based \cite{de2015pklot}. Counter-based methods are restricted to deployment in gated parking facilities, and they work by counting the number of vehicles that enter and exit the parking facility.  Sensor-based methods rely on physical detection sensors that are placed above or below parking lots, but are constrained by the significant capital costs of the large number of physical sensors required to cover large parking facilities \cite{de2015pklot}. Image-based methods rely on camera systems and are able to cover large outdoor or indoor parking lots when there are suitably high and unobstructed vantage points. Image-based methods also contain richer but less structured information than counter-based and sensor-based methods; for example, it is possible to identify specific vehicle characteristics from image-based methods but it is difficult to do so using counter-based or sensor-based methods.
	
	Huang et al \cite{huang2010hierarchical} further divide image-based methods into car-driven and space-driven methods. Car-driven methods primarily detects and tracks cars, and use car-detection results to quantify parking usage. Traditional object detection methods such as Viola et al \cite{viola2004robust} that rely on "simple" image representations and learning from training examples have been applied to identify vehicles in videos taken of parking lots by Lee et al \cite{lee2005automatic} and Huang et al \cite{huang2010hierarchical}. 
	
	\subsection{Space-driven methods}
	However, due to potential occlusions and perspective distortions of camera systems \cite{amato2017deep}, existing studies have instead focused on space-driven methods. Space-driven methods primarily observe changes in highlighted parking lots in an image frame. Past studies have used methods ranging from texture classifiers \cite{de2015pklot}, support vector machines \cite{bong2008integrated} and even recent deep learning-based methods \cite{amato2017deep,vu2018parking} to classify whether a parking space is occupied. These methods however rely on extensive, manual and relatively niche task labelling of the occupancy status of parking spots. For example, de Almeida et al \cite{de2015pklot} manually labelled 12,417 images of parking lots across multiple parking lots on the campus of Federal University of Parana (UFPR) and the Pontifical Catholic University of Parana (PUCPR) located in Curitiba, Brazil. 
	Besides the extensive effort required in obtaining image datasets and labelling them, the data collection process of parking spaces requires individual parking facilities to agree to data sharing and distribution. Fundamentally, space-based methods are not highly scalable, as they require extensive labelling and re-training of models for every distinct parking facility.
	
	\subsection{Car-driven methods}
	On the other hand, recent advancement in generic object detection through large-scale community projects led by organizations such as Microsoft \cite{lin2014microsoft} have allowed for access to large open datasets with more than 200,000 labelled images with more than 1.5 million object instances
	\cite{cocowebsite}. The labelled instances include labels for different motor vehicles, including trucks, buses, cars, and motorcycles, and are taken in a variety of contexts and image quality.
	
	In Figure \ref{fig:cocopklot}, we see that the extensive dataset labelled by de Almeida et al \cite{de2015pklot} consists of images taken from 3 spots, while the popular and open Common Objects in Context (COCO) dataset used for object detection takes images from a diverse range of perspectives.
	
	Past work in car-based parking quantification relies on traditional object detection \cite{huang2010hierarchical}. An example of such a method is proposed by Tsai et al \cite{tsai2007vehicle} that uses color to identify vehicle candidates, and trains a model that uses corners, edge maps and wavelet transform coefficients to verify candidates. While traditional computer vision techniques are able to achieve good levels of accuracy, they rely heavily on feature selection by researchers and hence may be sub-optimal. On the contrary, deep learning based computer vision techniques are able to automatically select and identify features in a hierarchical manner\cite{bengio2009learning,zhou2014learning}. 
	\section{Experiment and Data Collection}
\begin{figure}[H]
		\centering
		\resizebox{1.0\columnwidth}{!}{%
			\includegraphics[scale=0.5]{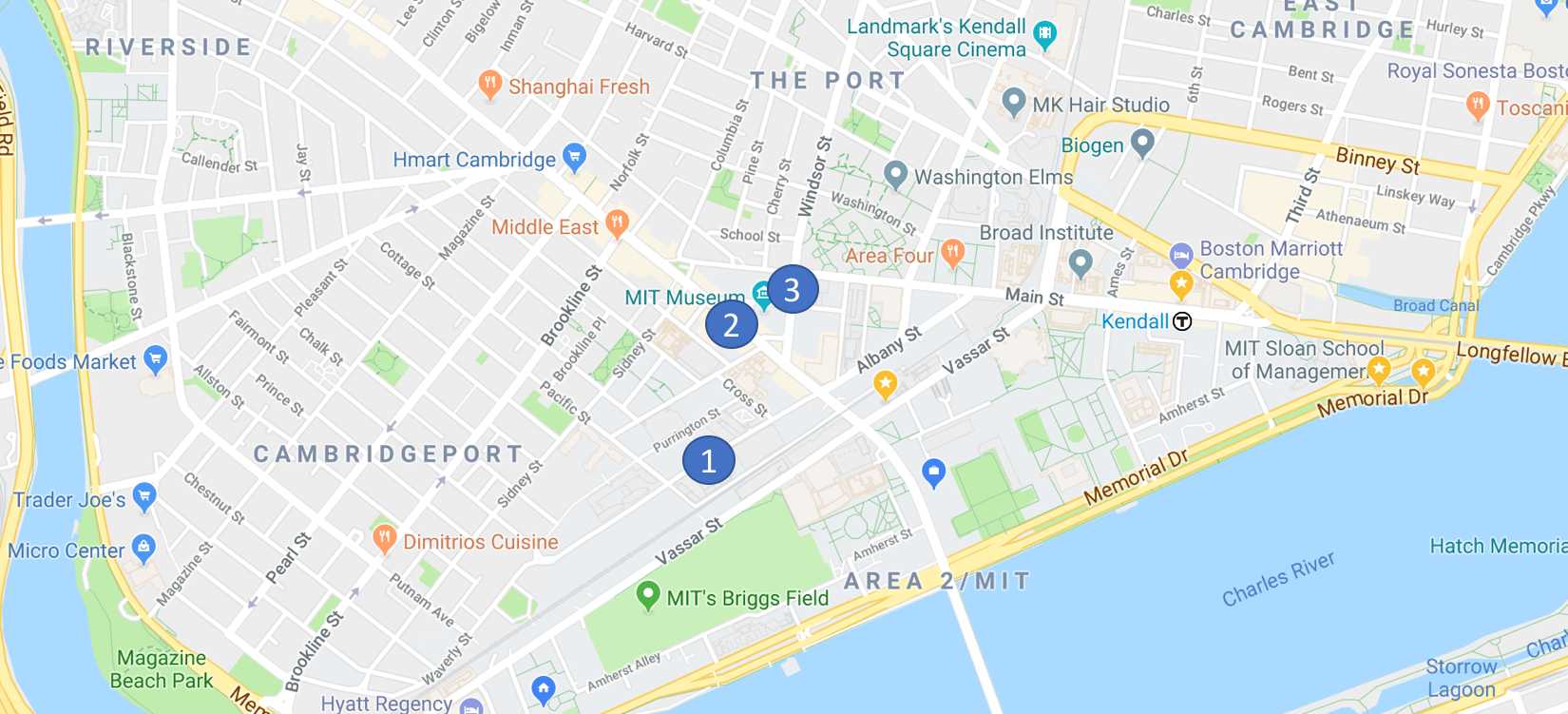}}
		\caption[Experiment sites]{Locations of street parking where video data was collected}
		\label{fig:sites}
	\end{figure}

	We collected 3 days of video footage of street parking around the MIT campus area in the City of Cambridge, USA. The locations of the studied parking lots can be seen in Figure \ref{fig:sites}. A summary of the sites is provided in Table \ref{table:sites}. 
	
	\begin{table}[h]
		\centering
		\resizebox{1.0\columnwidth}{!}{%
			\begin{tabular}{|c|cl|}
				\hline
				Site Name & Footage Length & Site Description \\
				& (minutes) & \\
				\hline
				\textit{Facilities} & 1696.5  & Four parking lots without lot boundaries.\\
				& & Camera mounted across street with small chance of occlusions. \\
				\hline
				\textit{IDC} & 3622.5 & Four parking lots with clear lot boundaries. \\
				& & Camera mounted at a high vantage point with no chance of occlusion. \\
				\hline
				\textit{Museum} & 3140.5 & Four parking lots with clear lot boundaries. Camera mounted at \\
				& & a low vantage point and across the street with high chance of occlusions. \\
				\hline
		\end{tabular}}
		\caption[Summary of experiment sites]{Descriptive summary of experiment sites}
		\label{table:sites}
	\end{table}
	The video footages were taken in the summer of 2017. The footage was collected from the duration of 04:00:00 to 23:59:59 therefore include footages when lighting conditions are not ideal. This is explained further later in the implementation challenges, and our validation results in Section V differentiates the complete results from results taken during visible and peak hours. We define visible and peak hours as the duration from 07:00:00 to 18:59:59, which overlaps with durations when the studied parking sites require payment, and when natural lighting is substantial at the sites.
	
	\section{Methodology}
	\subsection{Frame-wise Instance Segmentation}
	In this paper, we use the car-based method on images obtained. Unlike the space-based method, the car-based method depends on having an accurate and generalizable vehicle detector. The space-based method relies on classifying whether a specific parking space is occupied or not; this requires hand-labelling a specific parking facility and training a model that may not be generalizable to parking facilities other than the one that has been labelled.
	
	\begin{figure}
		\centering
		\resizebox{0.40\columnwidth}{!}{%
			\includegraphics[scale=0.5]{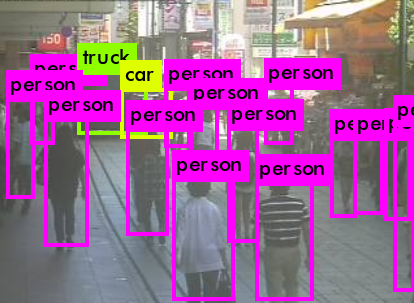}}
		\resizebox{0.54\columnwidth}{!}{%
			\includegraphics[scale=0.5]{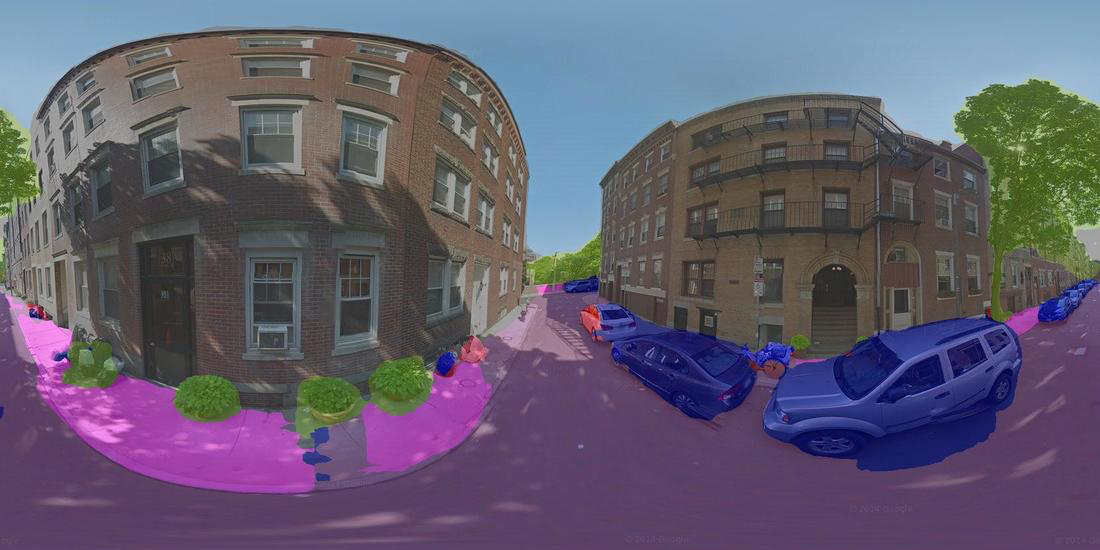}}\\
		\vspace{1mm}
		\resizebox{0.95\columnwidth}{!}{%
			\includegraphics[scale=0.5]{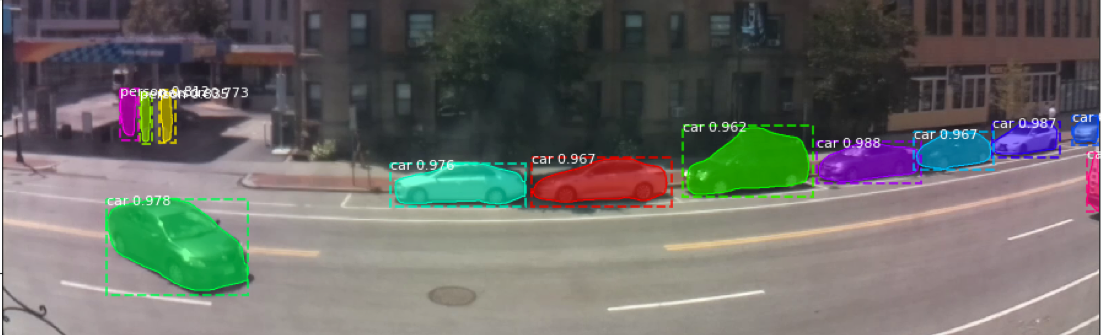}}
		\caption[Object detection, semantic segmentation, and instance segmentation]{Top left image shows the results of the object detection algorithm, which detects object classes and localizes them with a rectangular bounding box. Top right image shows the result of semantic segmentation algorithm, which labels every pixel in an image with an object class. Bottom image shows the result of the instance segmentation algorithm used in this study, which detects object classes of interests, localizes them with a bounding box, and also provides a pixel-level localization or mask of identified objects.}
		\label{fig:cvcomp}
	\end{figure}
	
	For the task of accurately quantifying parking space utilization, we use an instance segmentation algorithm as the baseline algorithm. Instance segmentation allows us to simultaneously identify individual instances of vehicles and precisely locate the boundaries of identified vehicle instances.
	
	There has been significant progress in the deep learning-based object detection and semantic segmentation literature that have enabled real-time and accurate performance. In the area of object detection, Ren et al \cite{ren2015faster} overcame a significant bottleneck in past object detection algorithms by replacing time-consuming traditional region proposal methods such as Selective Search \cite{uijlings2013selective} and EdgeBoxes \cite{zitnick2014edge} with a learnable and fast Region Proposal Network (RPN). For semantic segmentation, Long et al \cite{long2015fully} demonstrated that fully convolutional neural networks perform better than past neural net architectures with downsampling and upsampling. Yu et al later \cite{yu2015multi,yu2017dilated} introduced dilated convolutions that widen receptive fields of convolutional units.
	
	The algorithm that we employ for our purposes is based a Tensorflow implementation of He et al's \cite{he2017mask} Mask Region-based Convolutional Neural Network (Mask-RCNN). Based on Ren et al's \cite{ren2015faster} Faster-RCNN algorithm, Mask-RCNN adds a branch that predicts a mask or region-of-interest that serves as the pixel segmentation within the bounding boxes of each identified object instance. The simultaneous training and evaluation processes allow for fast training and real-time evaluation. Our Mask-RCNN implementation is trained using 2 NVIDIA GTX 1080Ti, and our training data is the COCO dataset, which has over 330 thousand training images, 1.5 million object instances and 80 distinct object categories, including cars, trucks, motorcycles and other motor vehicles. We used similar training parameters as reported by He et al \footnote{Training parameters are reported in \cite{he2017mask}, and also provided in the authors' Github repository: \href{https://github.com/facebookresearch/Detectron}{https://github.com/facebookresearch/Detectron}}, except that we used a decaying learning rate schedule with warm restarts that was introduced by Loshchilov et al \cite{loshsgdr2016}. Our Mask-RCNN implementation achieved an Average Precision (AP) score of 39.0 on the COCO test set, which is comparable to published Mask-RCNN metrics.
	
	\subsection{Parking Identification}
	\begin{figure}
		\centering
		\resizebox{\columnwidth}{!}{%
			\includegraphics[scale=0.5]{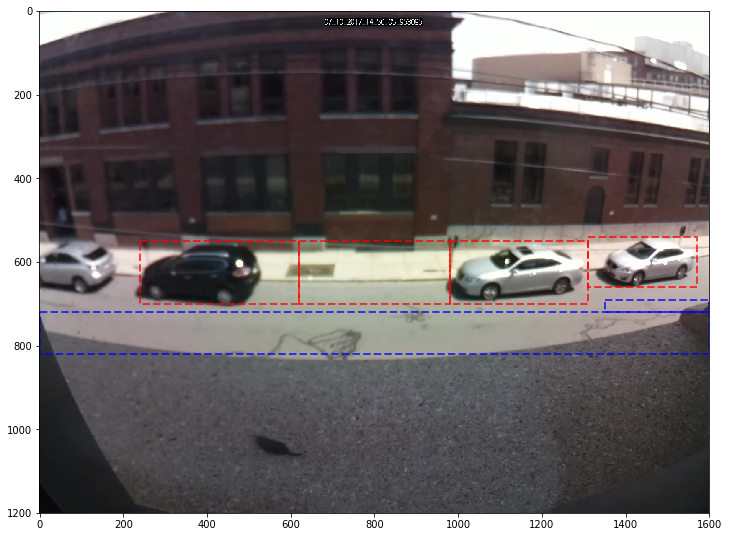}}
		\resizebox{\columnwidth}{!}{%
			\includegraphics[scale=0.5]{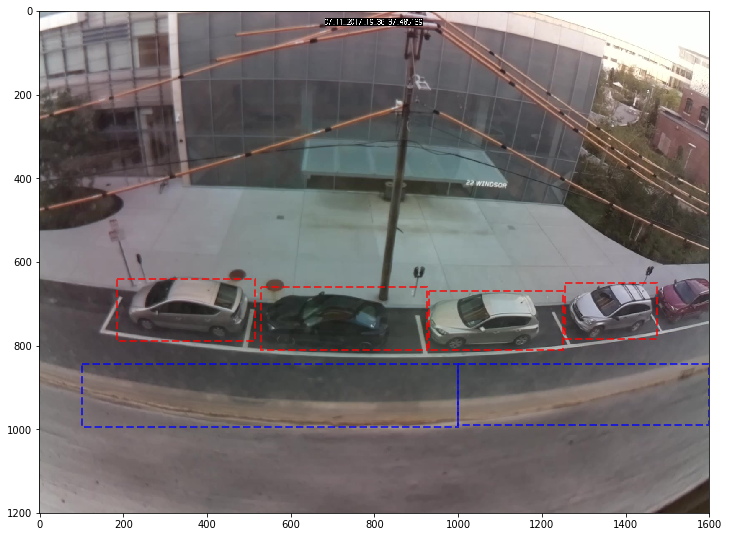}}\\
		\vspace{1mm}
		\resizebox{\columnwidth}{!}{%
			\includegraphics[scale=0.5]{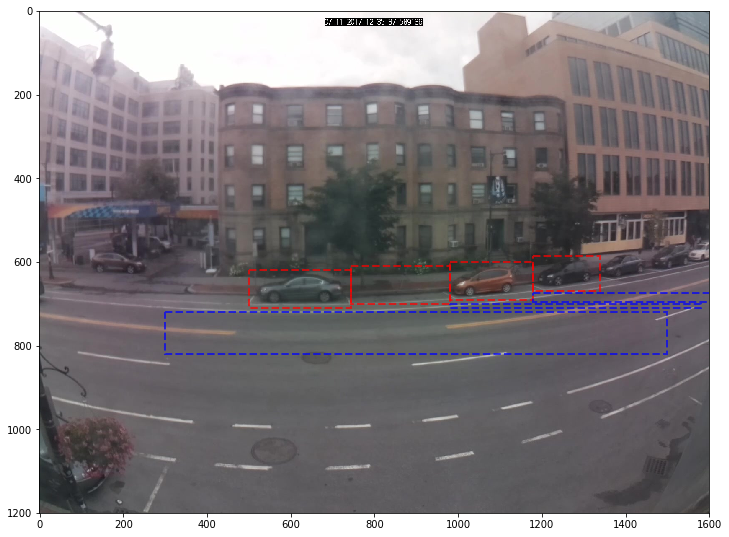}}
		\caption[Experiment sites with drawn parking lots and surrounding road areas]{Experiment sites with drawn parking lots and surrounding road areas. Top left image shows the \textit{Facilities} site, top right image shows the \textit{IDC} site, and bottom image shows the \textit{Museum} site.}
		\label{fig:parkingid}
	\end{figure}
	Parking lots are identified and surrounding road-areas are identified via hand-drawn labels. Using a simple area-based threshold, vehicles with a significant proportion of its area located in road-areas are determined to be either not parked or obstructions. Figure \ref{fig:parkingid} shows the identified parking lots in red and the surrounding road-areas in blue for all 3 studied sites. 
	
	\subsection{Implementation Challenges}
\begin{figure}[h]
		\centering
		\resizebox{1.0\columnwidth}{!}{%
			\includegraphics[scale=0.5]{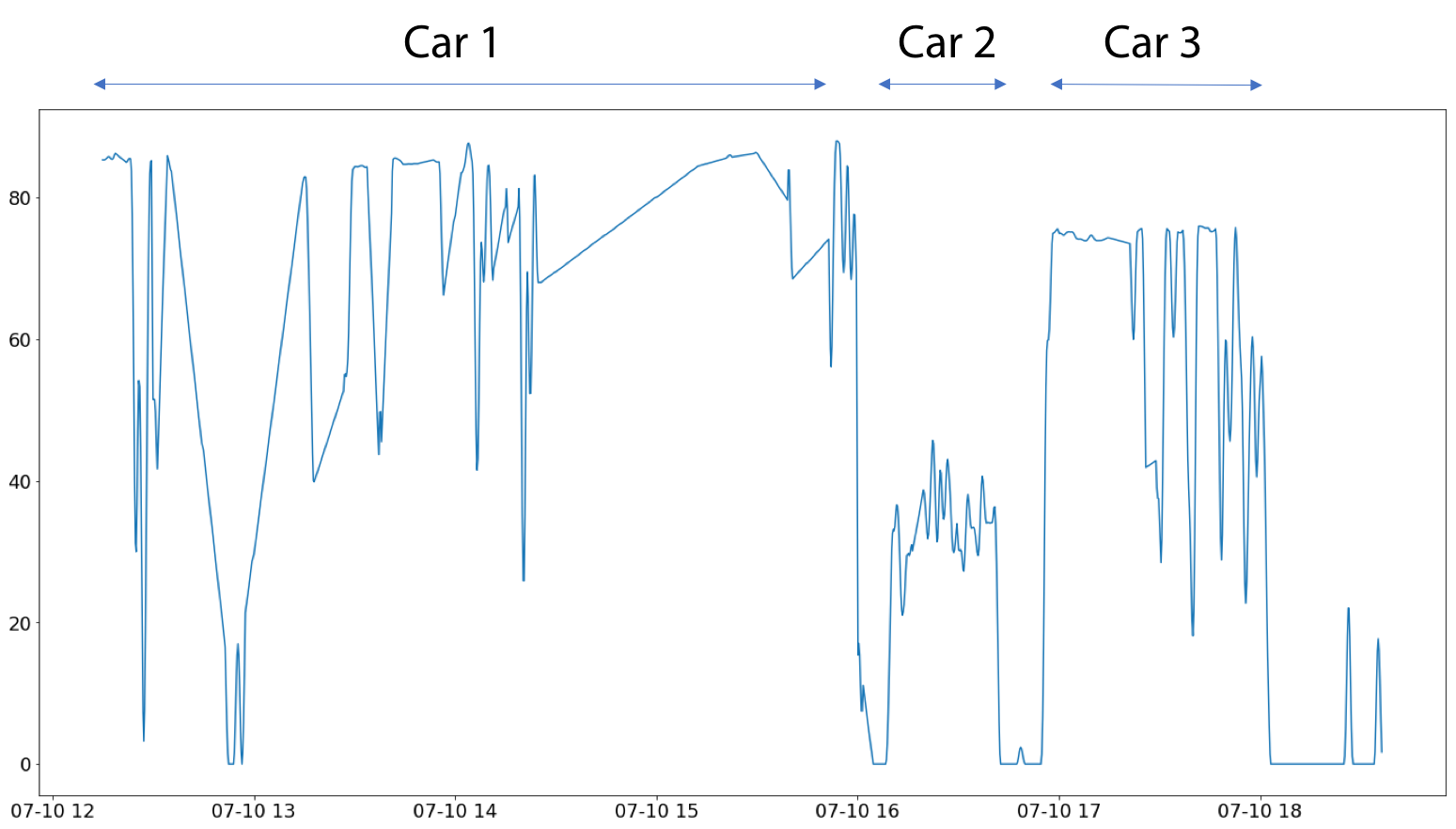}}
		\caption[Sample of noisy raw utilization measurements from \textit{Museum} site]{Sample of noisy raw utilization measurements from lot 3 of \textit{Museum} site. Actual car stay durations are drawn on top.}
		\label{fig:noisymuseumsample}
	\end{figure}

	We quantify parking utilization of lot $i$ and time $t$ as the ratio of the space utilization in the horizontal or x dimension and the horizontal space of lot $i$:
	\[
	\text{Utilization}_{i,t} = \frac{\text{Occupied horizontal space}_{i,t}}{\text{Horizontal space}_i}
	\]
	Using this measure, we find that a straightforward application of Mask-RCNN resulted in a noisy measurement of lot utilization. We directly applied Mask-RCNN to the recorded footages sampled at every 15 seconds, used the method described in 6.2.2 to identify vehicles that are parked, and applied the above definition to obtain parking utilization. For illustrative purposes, we focus on a particular duration of time at the \textit{Museum} site, and provide the utilization measurements and actual car stays during this duration for a single lot in Figure \ref{fig:noisymuseumsample}.
	
\begin{figure}[H]
		\centering
		\resizebox{1.0\columnwidth}{!}{%
			\includegraphics[scale=0.5]{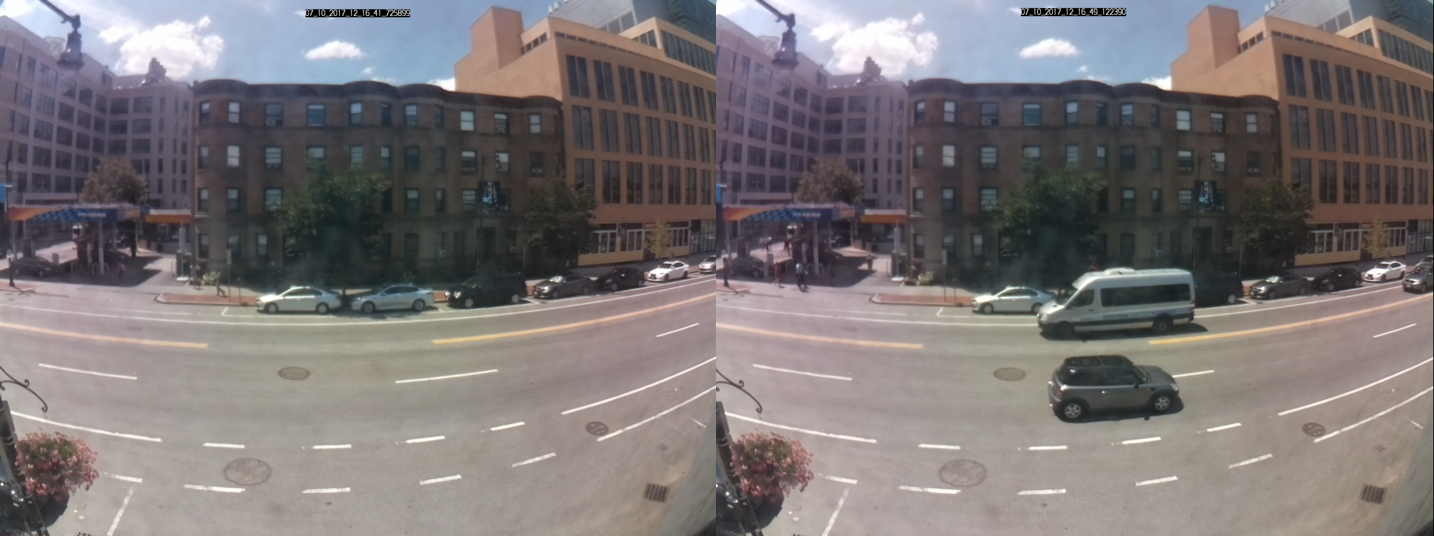}}
		\caption[Occlusion of street parking lots studied at the \textit{Museum} Site]{Unobstructed and occluded view of street parking lots at the \textit{Museum} Site}
		\label{fig:museumocclusion}
	\end{figure}

\begin{figure}[H]
		\centering
		\resizebox{1.0\columnwidth}{!}{%
			\includegraphics[scale=0.5]{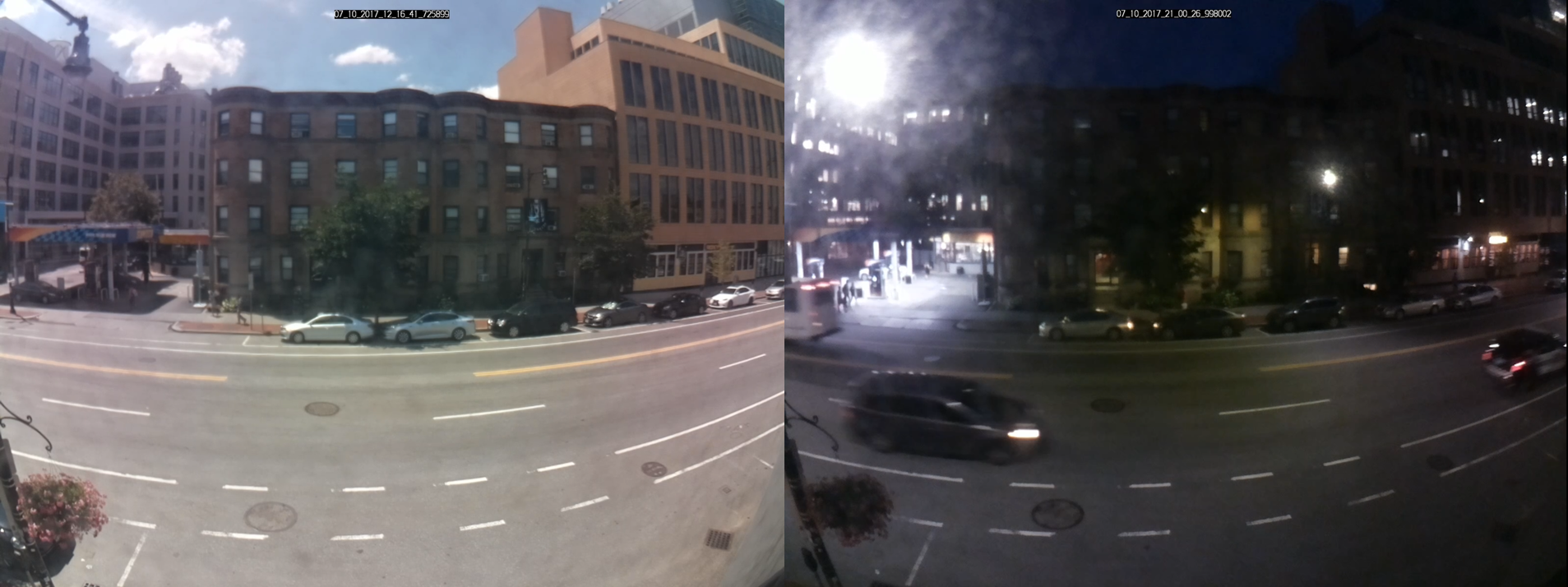}}
		\caption[Comparison of lighting conditions at street parking lots studied at the \textit{Museum} Site]{Unobstructed and occluded view of street parking lots at the \textit{Museum} Site}
		\label{fig:museumlighting}
	\end{figure}

	There are three factors that contribute to the noise seen in the utilization measurements:
	\begin{enumerate}
		\item \textbf{Occlusion}: A significant challenge for car-based methods is possible occlusion \cite{amato2017deep} of parking spots and observed vehicles. In the data collected, the occlusion problem is particularly severe for the \textit{Museum} site. Data at the \textit{Museum} site was collected via a camera mounted in the building across from the parking lots. This vantage point however is at a relatively low angle relative to the parking lots, hence data collected from this site often obstructed by vehicles in the street between the camera and the street parking lots of interest. Figure \ref{fig:museumocclusion} provides a side-by-side example of an unobstructed and occluded view of street parking lots at the \textit{Museum} site.
		\item \textbf{Weather and lighting conditions}: Figure \ref{fig:museumlighting} provides a side-by-side example of the \textit{Museum} site that illustrates the effect of changes in lighting conditions.
	\end{enumerate}
	\vspace{-7mm}
	\begin{enumerate}
		\setcounter{enumi}{2}
		\item \textbf{Random errors}: As we only consider detected cars if they exceed a certain threshold, idiosyncrasies in the video footage may result in random drops in detection, and also random false detections.
	\end{enumerate}
	\subsection{Smoothing Technique: Intelligent Car Tracking Filter} Signal processing techniques such as mode filters or low-pass frequency filters may not be effective in filtering out failures or noise in detection if they are not restricted to a particular (high) frequency domain. Furthermore, the use of such filters require calibration that is static. For example, the mode filter has a kernel size that would determine the length of maximum signal noise that it can handle.
	
	A major contribution of this paper is to introduce an intelligent car tracking filter. Instead of simply applying the mode filter on raw utilization values extracted using instance segmentation provided by Mask-RCNN, the intelligent car tracking filter maintains a memory of characterizing features of past cars detected, and compares across detected vehicles in the past to smooth detected car locations. 
	
	\begin{figure}[H]
		\centering
		\resizebox{1.0\columnwidth}{!}{%
			\includegraphics[scale=0.5]{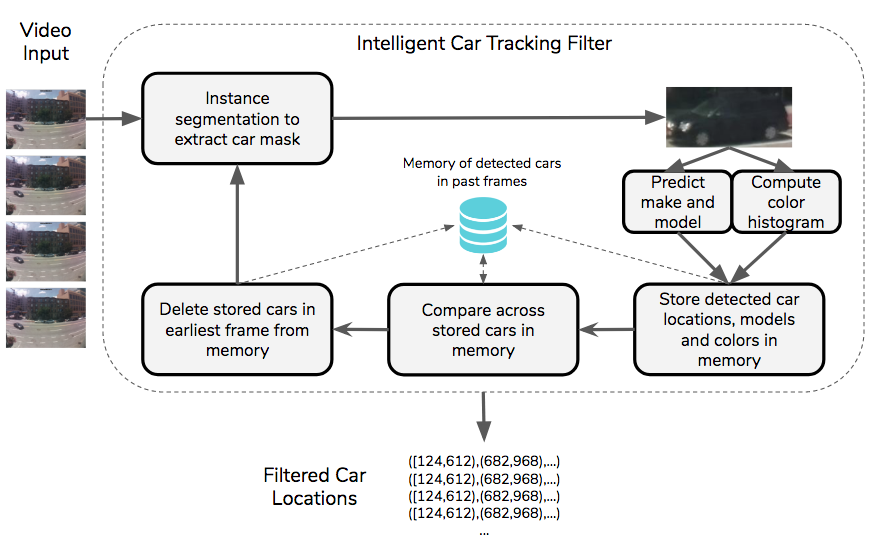}}
		\caption[Diagram of intelligent car tracking filter]{Diagram describing the intelligent car tracking filter}
		\label{fig:filterdiagram}
	\end{figure}
	
	\noindent As illustrated in Figure \ref{fig:filterdiagram}, the filter repeats the following steps: \begin{enumerate}
		\item Run instance segmentation on sampled images to extract car masks
		\item Vehicle features are extracted from detected vehicles\\
		(a) A car classification model is applied on extracted vehicle masks to obtain a vector of features\\
		(b) Color histograms of the two vehicles are computed as a feature vector
		\item Car model feature vector and color histogram feature vectors are stored in memory
		\item Past identified cars are compared and matched based on car model, color histogram feature vectors, and locations of cars in memory, to filter car locations
		\item Cars from earliest frame is deleted
	\end{enumerate}
	\noindent For step 2 (a), we train the car classification model using the Stanford Cars dataset, which contains 196 different make and models of cars in a dataset of 16,185 images \cite{cardataset}. We use the ResNet-50 architecture for the car classification model due to its residual network structure that allows for effective learning for deep neural networks \cite{he2016deep}. By running the trained car classification model on the detected car instances, we obtain a feature vector $X_c\in\mathbb{R}^K,X_c\in(0,1)^K,\;K=196$. For step 2 (b), we obtain a feature vector $X_h\in\mathbb{R}^M,X_c\in(0,1)^M,\;M=24$ that characterizes the color histogram of the extracted vehicle mask.
	
	We consider the case that the intelligent car tracking filter is applied to a video input streamed at a consistent frame rate of 1 frame every $S$ seconds, and the memory of the filter keeps track of all cars detected in the past $n$ frames. Without loss of generality, we assume that $n$ is odd. Consider that at time $t$, we should optimally infer the locations of cars at time $t-\text{round}(\frac{n}{2})\cdot S$, or $\text{round}(\frac{n}{2})$ frames ago.
	
\begin{figure}[h]
		\centering
		\resizebox{1.0\columnwidth}{!}{%
			\includegraphics[scale=0.5]{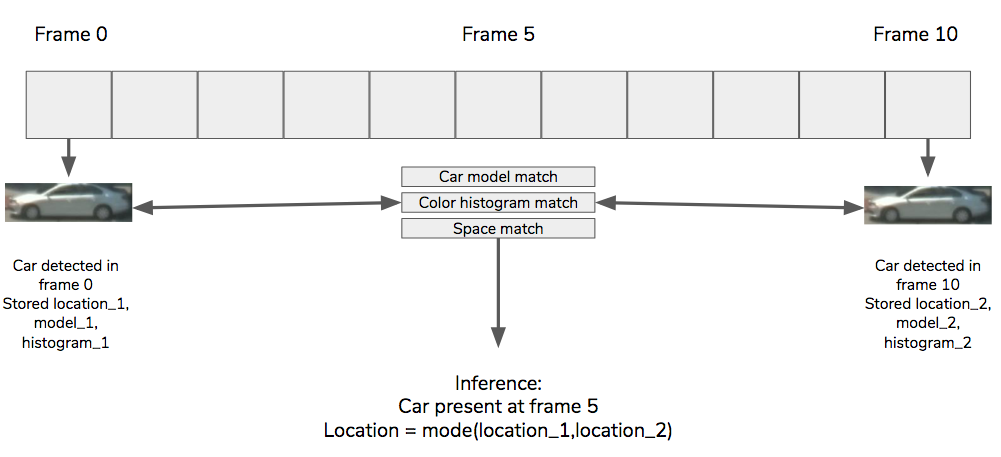}}
		\caption[Maximum occlusion that can be handled by filter]{Diagram describing the maximum occlusion scenario that the intelligent filter can handle}
		\label{fig:extremecase}
	\end{figure}

\begin{figure}[h]
		\centering
		\resizebox{1.0\columnwidth}{!}{%
			\includegraphics[scale=0.5]{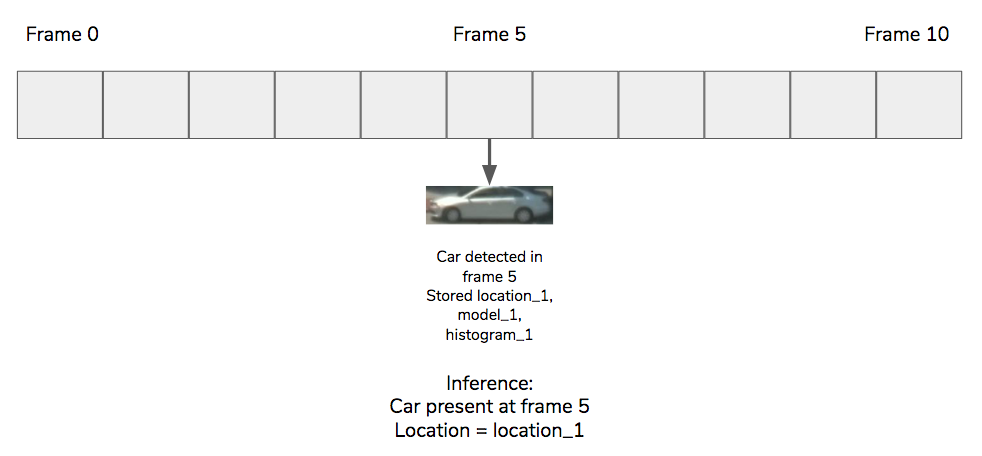}}
		\caption[Single detection at frame $\text{round}(\frac{n}{2})$]{Diagram describing the single detection case}
		\label{fig:singlecase}
	\end{figure}

\begin{figure}[h]
		\centering
		\resizebox{1.0\columnwidth}{!}{%
			\includegraphics[scale=0.5]{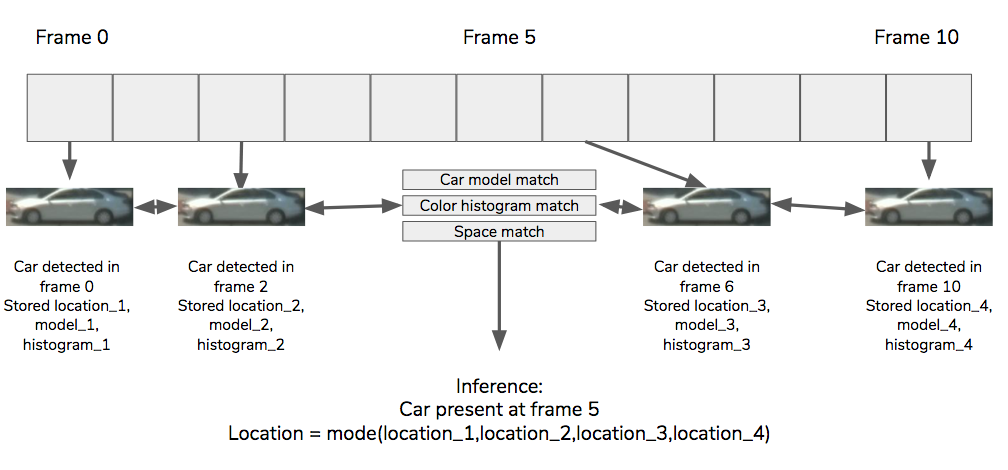}}
		\caption[Typical scenario for matching cars]{Diagram describing the typical scenario when the matched car is detected multiple times}
		\label{fig:usualcase}
	\end{figure}

	Figure \ref{fig:extremecase} describes this extreme case with a scenario where $n=11$. In general, with a memory of $n$ frames and at present time $t$, inferring the locations of cars in the past at time $t-\text{round}(\frac{n}{2})\cdot S$, or $\text{round}(\frac{n}{2})$ frames ago would handle the maximum duration of occlusion. Step 4 does this by matching cars in its memory and inferring that the car is present at frame $\text{round}(\frac{n}{2})$ if  (1) the matched car is found before and after frame $\text{round}(\frac{n}{2})$, and/or (2) the car is detected at frame $\text{round}(\frac{n}{2})$. Figure \ref{fig:singlecase} illustrates the scenario when the car is only detected at frame $\text{round}(\frac{n}{2})$, while Figure \ref{fig:usualcase} describes the typical scenario when the matched car is detected multiple times.
	\vspace{-3mm}
	\\
	A pair of cars $A$ and $B$ have model feature vectors $X_c$, color histograms $X_h$ and horizontal pixel locations of the detected cars $X_l$, where $X_l\in\mathbb{R}^2$. We consider this pair of cars as matched if $||X_c^A-X_c^B||_1<T_c$, $D_B(X_h^A-X_h^B)<T_b$ and $||X_l^A-X_l^B||_1<T_l$, where $T_c,T_b,T_l$ are threshold levels for matches calibrated by manually looking at pairs of extracted vehicle masks, and $D_B$ is the Bhattacharyya distance function that measures similarity between two distributions \cite{bhattacharyya1943measure}. Once the detected car instances are matched and the filter infers that the car is present at frame $\text{round}(\frac{n}{2})$, the system returns the mode of all detected $X_l$.
	
\begin{figure}
		\centering
		\resizebox{1.0\columnwidth}{!}{%
			\includegraphics[scale=0.5]{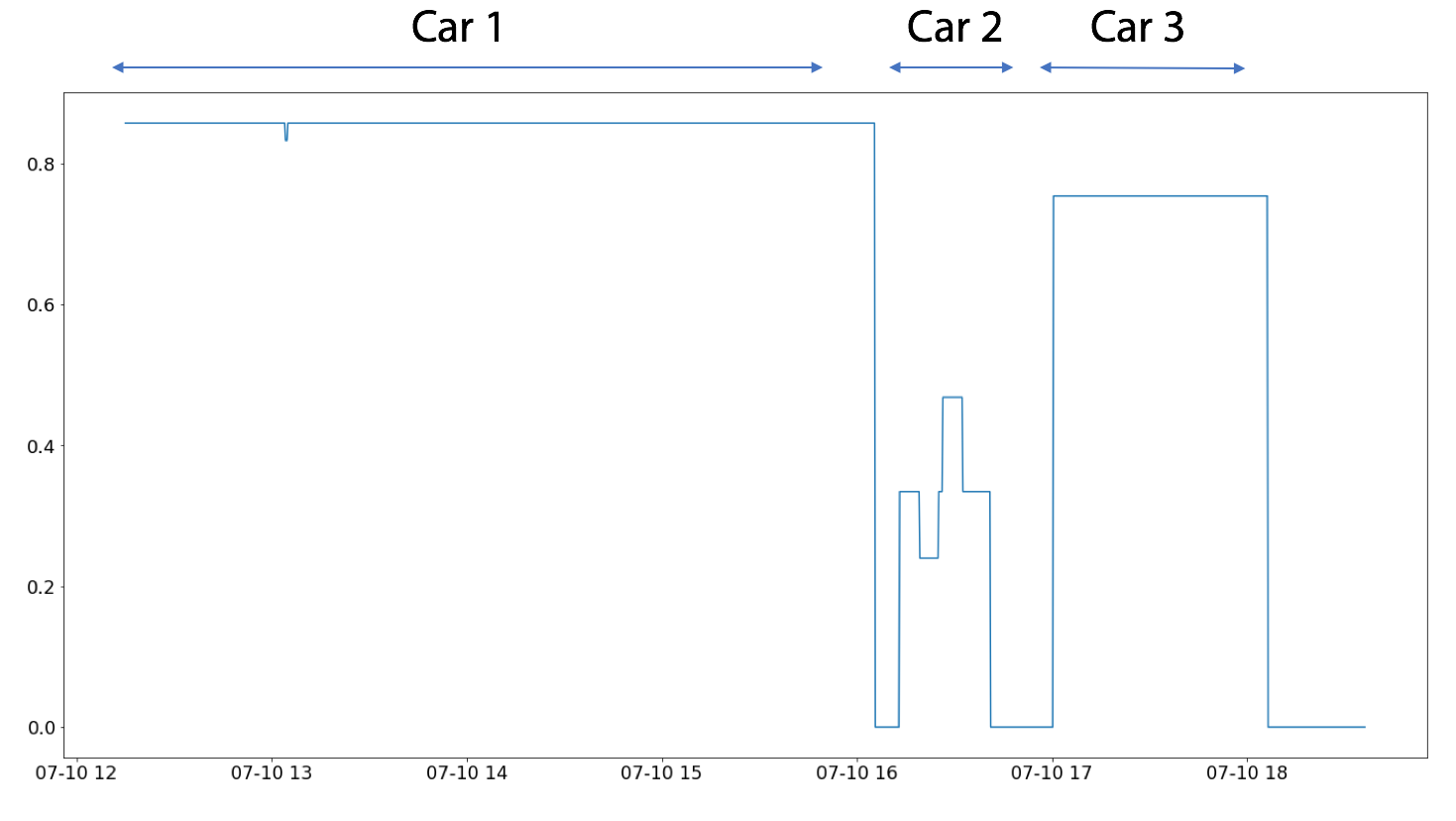}}
		\caption[Intelligent filter applied to sample of utilization measurements from \textit{Museum} site]{Intelligent filter applied to sample of utilization measurements from \textit{Museum} site}
		\label{fig:filterdemo}
	\end{figure}

	In summary, the intelligent car tracking filter acts as a mode filter with a dynamic kernel that adjusts to matched vehicle instances in order to smooth the identified car locations. As a brief visual illustration of its effectiveness, we applied the intelligent car tracking filter on the same duration for a single lot in the \textit{Museum} site as seen in Figure \ref{fig:noisymuseumsample}, and provide the results in Figure \ref{fig:filterdemo}.
	\section{Validation Data and Metrics}
	
	With actual deployment in a smart parking enforcement or payment in mind, we evaluate the filter based on its detection, spatial and time accuracy, as well as processing speed. To validate these metrics, we went through the recorded videos and manually labelled randomly selected frames or for randomly selected vehicles. Table \ref{table:labels} describes the validation datasets that we constructed.
	
	\begin{table}[h]
		\centering
		\resizebox{1.0\columnwidth}{!}{%
			\begin{tabular}{|c|cl|}
				\hline
				Validation Metrics & Dataset Size & Label Description \\
				\hline
				Detection and & 138 frames & Randomly selected 138 frames\\
				Spatial Accuracy & 309 vehicles & across all 3 sites and labelled bounding \\
				& & boxes for parked vehicle instances  \\
				\hline
				Time & 40 vehicles & Randomly selected 40 vehicles across  \\
				Accuracy & 3410 mins & all 3 sites and recorded the time that the\\
				& & vehicle entered and left its parking lot\\
				\hline
		\end{tabular}}
		\caption[Summary of validation datasets]{Descriptive summary of validation datasets}
		\label{table:labels}
	\end{table}
	
	For the first dataset in Table \ref{table:labels}, we only used frames where all parking lots of interest are unobstructed. Furthermore, both datasets were obtained via random sampling from the entire timeframe of 04:00:00 to 23:59:59, which includes periods of poor visibility. As mentioned earlier, we further used a subset of the validation set where all data was restricted to the duration between 07:00:00 to 18:59:59. The size of the first dataset reduces to 113 frames and 253 vehicles, while the size of the second dataset reduces to 30 vehicles with total duration of 1626 mins of validation footage.
	
	\subsection{Detection Accuracy}
	
	Using the first dataset described in Table \ref{table:labels}, we use a simple ratio of vehicles that are detected and labelled to be present (denoted as $TP$), over the sum of vehicles that are detected and labelled to be present, vehicles that are detected but not labelled (denoted as $FP$), and vehicles that are not detected and labelled, (denoted as $FN$). Denoting $TP_i,FP_i,FN_i$ as the evaluation metrics for the $i$th frame in the validation dataset, we first sum these metrics across all 150 validation frames before taking the ratio:
	
	\[
	\text{Detection Accuracy} = \frac{\sum\limits_{i=1}^{150}TP_i}{
		\sum\limits_{i=1}^{150}(TP_i+FP_i+FN_i)}
	\]
	
	\subsection{Spatial Accuracy}
	
	We further use the first dataset described in Table \ref{table:labels} to validate the spatial accuracy of this method. For each $j$th detected vehicle instance, we have a manually labelled bounding box and a detected vehicle mask. We extract the left and right horizontal pixel coordinates of the bounding box (denoted as $X_{l,\text{true}}^j$ and $X_{r,\text{true}}^j$), and also the leftmost and rightmost horizontal pixel coordinates of the vehicle mask (denoted as $X_{l,\text{output}}^j$ and $X_{r,\text{output}}^j$). Denoting $N_{TP}=\sum\limits_{i=1}^{150} TP_i$ as the total number of all detected and labelled vehicle instances, we define spatial accuracy as the average ratio of horizontal area of intersection between box and mask over the horizontal area of union between box and mask. Note that the following definition follows the convention where $X_{r,\text{true}}^j\geq X_{l,\text{true}}^j$ and $X_{r,\text{output}}^j\geq X_{l,\text{output}}^j$.
	
	\begin{equation*}
	\resizebox{\columnwidth}{!}{%
		$\text{Spatial Accuracy} = \frac{1}{N_{TP}}\cdot\sum\limits_{j=1}^{N_{TP}}\frac{\min(X_{r,\text{true}}^j,X_{r,\text{output}}^j)-\max(X_{l,\text{true}}^j,X_{l,\text{output}}^j)}{\max(X_{r,\text{true}}^j,X_{r,\text{output}}^j) - \min(X_{l,\text{true}}^j,X_{l,\text{output}}^j)}
		$}
	\end{equation*}
	
	\subsection{Time Accuracy}
	
	We use the second dataset described in Table \ref{table:labels} to evaluate the time accuracy of the filter. The time accuracy metric measures the ratio of detected occupancy of a particular lot over the actual occupancy of a particular lot. Denoting the number of frames when a vehicle is detected as being in the lot for the $i$th validated vehicle as $F_{i,\text{output}}$, and the total number of frames when the $i$th validated vehicle as actually labelled being parked in the lot as $F_{i,\text{true}}$, we define the first time accuracy metric as:
	\small
	\[
	\text{Time Accuracy} = \frac{\sum\limits_{i=1}^{42} F_{i,\text{output}}}{\sum\limits_{i=1}^{42}F_{i,\text{true}}}
	\]
	\normalsize
	The San Francisco Metropolitan Transport Authority defines this metric as the occupancy accuracy metric, and uses this metric to evaluate potential vendors for SFPark, a project aimed at managing demand for parking spaces in San Francisco \cite{sfparkreport}. 
	\subsection{Processing Efficiency}
	We measure the average processing time per sampled frame as the benchmark for processing efficiency, and also the total processing time for the entire duration of videos taken at all 3 sites.
	\section{Validation Results}
	\subsection{Accuracy Results: Pure Detection and Memory Filter}
	\begin{table}[h]
		\centering
		\begin{tabular}{|c|ccc|}
			\hline
			Memory Size & Detection & Spatial & Time\\
			(\# of frames) & Accuracy& Accuracy & $\text{Accuracy}$\\
			& (\%) & (\%) & (\%) \\
			\hline
			Mask-RCNN Only & 89.2 & 92.1 & 64.9  \\
			5 & 90.2 & 91.2 & 71.3\\
			25 & 94.0 & 91.0 & 80.0 \\
			50 & 94.6 & 88.9 & 83.2 \\
			100 & 93.5 & 88.7 & 87.5 \\
			150 & 93.0 & 86.1 & 87.9 \\
			
			\hline
		\end{tabular}
		\caption[Accuracy results from validation datasets sampled from full timeframe]{Accuracy results from validation datasets sampled from full timeframe}
		\label{table:fullresults}
	\end{table}
	\begin{table}[h]
		\centering
		\begin{tabular}{|c|ccc|}
			\hline
			Memory Size & Detection & Spatial & Time \\
			(\# of frames) & Accuracy& Accuracy & $\text{Accuracy}$\\
			& (\%) & (\%) & (\%) \\
			\hline
			Mask-RCNN Only & 90.0 & 92.2 & 72.2 \\
			5 & 89.7 & 91.6 & 78.3 \\
			25 & 93.1 & 91.6 & 88.2\\
			50 & 93.5 & 90.0 & 91.2 \\
			100 & 92.2 & 89.4 & 95.8 \\
			150 & 91.6 & 88.7 & 96.1\\
			
			\hline
		\end{tabular}
		\caption[Accuracy results from validation datasets sampled from 07:00:00 to 18:59:59]{Accuracy results from validation datasets sampled from 07:00:00 to 18:59:59}
		\label{table:peakresults}
	\end{table}
	We compare the results of simply running the Mask-RCNN instance segmentation algorithm (Mask-RCNN Only in Tables \ref{table:fullresults} and \ref{table:peakresults}) against applying our filter on top of Mask-RCNN with different memory sizes on the video footages sampled at 1 frame every 15 seconds. Validation results in both the full timeframe (Table \ref{table:fullresults}) and the peak timeframe (Table \ref{table:peakresults}) demonstrate that the filter significantly increases $\text{Time Accuracy}$, and slightly increases Detection Accuracy. The significant improvements in the Time Accuracy metric can be attributed to the inferences that the intelligent filter are able to make. 
	
	Comparing results between in the peak and full timeframes, we see that accuracy is generally higher in the peak timeframes. This reflects the sensitivity of image-based methods to poor lighting during nighttime, especially since the \textit{Museum} and \text{Facilities} sites are not well-lit in the night.
	\subsection{Accuracy Results: Comparison to Industry Benchmarks}
	In 2014, SFPark evaluated trial parking sensor systems including image recognition systems, radar sensors and infrared cameras. One key metric that SFPark tracked was the occupancy accuracy metric that is defined identically with $\text{Time Accuracy}$ \cite{sfparkdataguide}, and charts\footnote{Pages 12 and 18 have charts that suggest that the validation was conducted during parking meter operation hours that are contained within our peak timeframe definition \cite{sfparkreport}.} in the SFPark evaluation report suggests that the evaluation was conducted in daytime. The results show that our intelligent filter significantly outperforms the industry benchmark image method provided by Cysen. Furthermore, the image recognition benchmark has similar performance to simply applying Mask-RCNN during the peak time period (see Detection Only - Peak in Table \ref{table:industrycomp}), suggesting that existing commercial vendors utilize an image-based rather than a video-based approach. In addition, the performance of the intelligent filter in the peak time period (see Intelligent Filter - Peak in Table \ref{table:industrycomp}) is comparable to the performances of the best sensor systems evaluated by SFPark. 
	\begin{table}[h]
		\centering
		\begin{tabular}{|c|c|}
			\hline
			Sensor System (Vendor) & $\text{Time Accuracy}$ \\
			\hline
			Radar/Magnetometer (Fybr) & 78\% \\
			Radar (Sensys) & 98\% \\
			Infrared (CPT) & 92\% \\
			Image Recognition (Cysen) & 77\% \\
			Magnetometer (StreetSmart) & 81\% \\
			\hline
			Detection Only - Full & 65\% \\
			Detection Only - Peak & 72\% \\
			Intelligent Filter - Full & 88\% \\
			Intelligent Filter - Peak & 96\%\\
			\hline
		\end{tabular}
		\caption[Occupancy accuracy comparison with industry standards]{Comparison of occupancy accuracy of intelligent filter with industry standards. Top half of table contains industry benchmarks while bottom half shows our validation results}
		\label{table:industrycomp}
	\end{table}
	\vspace{-8mm}
	\subsection{Processing Speed}
	We used a single computer with an Intel i7-7700K CPU, 16GB of RAM and a NVIDIA GTX-1080Ti GPU that has a retail market price of \$1,500 to evaluate the average processing time and total processing time. Table \ref{table:speedcomp} shows the processing speed in terms of average processing time per frame, and total processing time for the entire duration of our recorded footage..
	\begin{table}[H]
		\centering
		\begin{tabular}{|c|cc|}
			\hline
			Memory Size & Average Processing Time & Total Processing Time\\
			(\# of frames) & Per Frame (seconds) & (seconds) \\
			\hline
			Mask-RCNN Only & 0.399 & 13501\\
			5 & 0.413 & 13975\\
			25 & 0.435 & 14720 \\
			50 & 0.469 & 15870 \\
			100 & 0.556 & 18814 \\
			150 & 0.673 & 22773 \\
			\hline
		\end{tabular}
		\caption[Processing speed comparison]{Comparison of processing speed of filter with different memory sizes}
		\label{table:speedcomp}
	\end{table}
	\section{Conclusion}
	\begin{figure*}[h]
		\centering
		\resizebox{1.0\textwidth}{!}{%
			\includegraphics[scale=0.5]{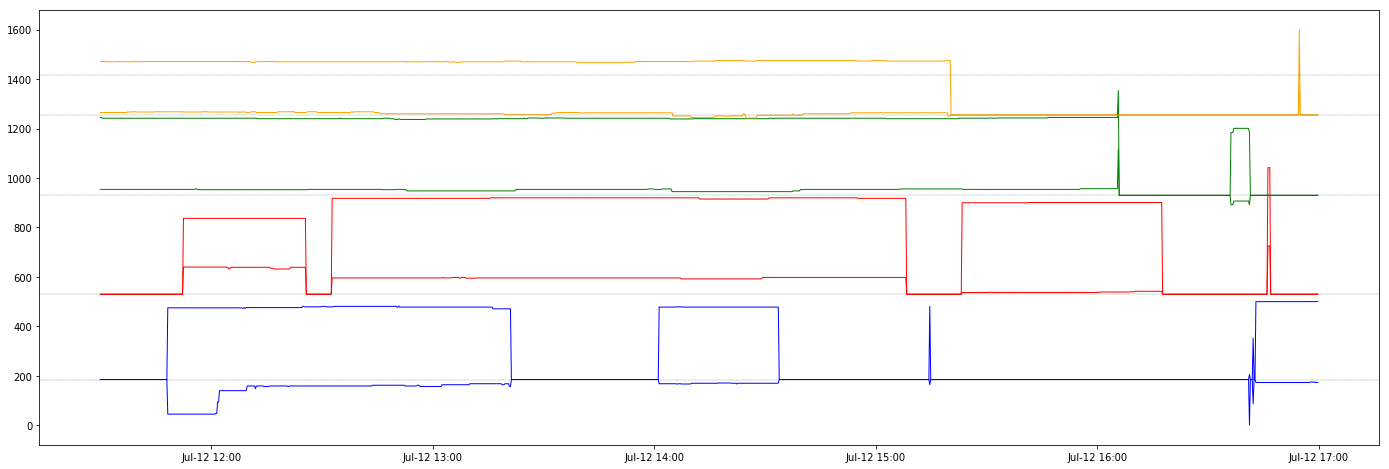}}
		\caption[Vehicle location information generated from intelligent filter]{Vehicle location information for the \textit{Museum} site generated from intelligent filter. Grey lines mark boundaries between parking lots. Blue lines, red lines, green lines, and yellow lines mark the boundaries of car parked in lot 0, 1, 2, and 3 respectively.}
		\label{fig:ecg}
	\end{figure*}
	The validation results demonstrate that our method significantly improves accuracy by treating the parking measurement problem as a \textit{video} problem rather than an \textit{image} problem. By combining information from video frames before and after, our system is able to better infer vehicle occupancy as compared to pure image-based methods. Furthermore, evaluation results by SFMTA supports the claim that our system is comparable in performance to the advanced commercial systems that employ more expensive sensors. For further verification of our system's relative performance to other methods, future studies should compare different sensor systems on identical parking sites and at identical periods of time.
	
	\subsection{Feasibility as a city-wide system}
	
	The fast average processing time per frame shows that our system can be financially feasible in a real-world deployment. Using a memory size of 100 frames, and a video sample rate of 1 frame every 15 seconds, the computer that we used for evaluation would be able to handle up to 89 parking lots. This translates to a cost of around \$17 per parking lot for processing ability, and we estimate that a camera, a single board computer and other associated costs would cost a further \$15 per parking lot. The eventual cost of around \$32 per parking lot is highly competitive as compared to existing sensors such as ground-based parking sensors that costs up to \$200 per parking lot\footnote{For example, PNI Corp prices its PlacePod ground sensors at \$200 per parking sensor. Price obtained from \url{ https://www.pnicorp.com/product-category/smart-parking/}.}. In demonstrating that our method is accurate and competitive at a per parking lot level basis, our paper opens up the opportunity for further research into the scalability of a camera and video based street parking monitoring system at a city-wide scale, especially in comparison to existing projects and technologies such as SFPark.
	
	\subsection{Tradeoff between immediacy and accuracy}
	
	An important qualification is that a feature of our implementation of the intelligent filter is that the filtered car location information is almost real-time. This is due to the inference that the system needs to make, which utilizes both information from "future" and "past" frames. Depending on the system's application in parking enforcement, quantification or real-time information to drivers, the system can make inferences about frames closer to the present time, $t$, and incur losses in accuracy. This change can be done by switching the inference frame from the current $t-\text{round}(\frac{n}{2})$ frame to a frame that is closer to the present time $t$.
	
	\subsection{Generalizability to other sites}
	
	We experienced little difficulty in extending our system to the three sites that we studied. Unlike space-based methods that require labelling and training for every distinct parking facility, we only need to mark out parking lot boundaries and surrounding road areas once to configure our system for a new parking facility. Labelling was only required to validate our results.
	
	\subsection{Information beyond binary occupancy}
	Our system can provide richer information than traditional binary occupancy information, especially in detecting space-based information and car make and color information. Figure \ref{fig:ecg} plots the left and right boundaries of filtered car locations for the \textit{Museum} site. Furthermore, our video-based method is also capable of recognizing specific type of vehicles, including conventional taxis and delivery trucks. Hence, our method provides the technical foundation for richer ways to understand curbside and parking occupancy that exceeds beyond traditional binary parking statistics.

	\section*{Acknowledgment}
	\noindent The authors would like to thank Philips Lighting for supporting this project. In addition, the authors would also like to thank Allianz,  Amsterdam Institute for Advanced Metropolitan Solutions, Brose, Cisco, Ericsson, Fraunhofer Institute, Liberty Mutual Institute, Kuwait-MIT Center for Natural Resources and the Environment, Shenzhen, Singapore- MIT Alliance for Research and Technology (SMART), UBER, Victoria State Government, Volkswagen Group America, and all the members of the MIT Senseable City Lab Consortium for supporting the lab's research.

	\ifCLASSOPTIONcaptionsoff
	\newpage
	\fi

	
	
	\bibliographystyle{IEEEtran}
	\bibliography{bib}
	%
	
	%







\end{document}